\documentclass[letterpaper, 10 pt, conference]{ieeeconf}  

\IEEEoverridecommandlockouts                              

\usepackage{graphicx}
\usepackage{graphics} 
\usepackage{amsmath} 
\usepackage{amssymb}  
\usepackage{multirow}
\usepackage{comment}
\usepackage{bm}
\usepackage{amsbsy}
\usepackage{xcolor}
\usepackage{hyperref}
\let\labelindent\relax
\usepackage{enumitem}

\title{\LARGE \bf
End-to-End Learning to Grasp via Sampling from Object Point Clouds$^*$
}

\author{Antonio Alliegro$^{1}$, Martin Rudorfer$^{2}$, Fabio Frattin$^{1}$, Ale\v{s} Leonardis$^{2}$ and Tatiana Tommasi$^{1}$%
\thanks{$^*$This work was partially supported by the CHIST-ERA BURG project under EPSRC grant no. EP/S032487/1.}%
\thanks{$^{1}$A.~Alliegro, F.~Frattin and T.~Tommasi are with the DAUIN Department at Politecnico di Torino. A.~Alliegro and T.~Tommasi are also affiliated with the Italian Institute of Technology, Italy.  
        {\tt\small \{antonio.alliegro,} {\tt\small fabio.frattin,tatiana.tommasi\}@polito.it}}%
\thanks{$^{2}$M.~Rudorfer and A.~Leonardis are with the School of Computer Science at the University of Birmingham, UK.
        {\tt\small \{m.rudorfer, a.leonardis\}@bham.ac.uk}}%
}

\newcommand{\bx}{\boldsymbol{x}}

\newcommand{\bc}{\boldsymbol{c}}

\newcommand{\bp}{\boldsymbol{p}}
\newcommand{\bq}{\boldsymbol{q}}

\newcommand{\bu}{\boldsymbol{u}}

\newcommand{\by}{\boldsymbol{y}}

\newcommand{\bg}{\boldsymbol{g}}

\newcommand{\br}{\boldsymbol{r}}

\newcommand{\sC}{\mathcal{C}}

\newcommand{\sP}{\mathcal{P}}
\newcommand{\sF}{\mathcal{F}}
\newcommand{\sQ}{\mathcal{Q}}
\newcommand{\sN}{\mathcal{N}}
\newcommand{\sG}{\mathcal{G}}

\newcommand{\sL}{\mathcal{L}}

\newcommand{\field}[1]{\mathbb{#1}}

\newcommand{\R}{\field{R}}

\newcommand*{\deco}{DeCo\@\xspace}

\newcommand{\tat}[1]{{\color{black}#1}}

\usepackage{xspace}

\makeatletter
\DeclareRobustCommand\onedot{\futurelet\@let@token\@onedot}
\def\@onedot{\ifx\@let@token.\else.\null\fi\xspace}

\def\ie{\textit{i.e}\onedot}

\makeatother

\usepackage{accents}

\usepackage{xcolor}
\definecolor{darkpink}{rgb}{0.91, 0.33, 0.5}
\definecolor{aquamarine}{rgb}{0.5, 1.0, 0.83}

\makeatother

\usepackage{ulem}

\begin{document}

\maketitle
\thispagestyle{empty}
\pagestyle{empty}


\begin{abstract}
The ability to grasp objects is an essential skill that enables many robotic manipulation tasks.
Recent works have studied point cloud-based methods for object grasping by starting from simulated datasets and have shown promising performance in real-world scenarios.
Nevertheless, many of them still rely on ad-hoc geometric heuristics to generate grasp candidates, which fail to generalize to objects with significantly different shapes with respect to those observed during training. 
Several approaches exploit complex multi-stage learning strategies and local neighborhood feature extraction while ignoring semantic global information.
Furthermore, they are inefficient in terms of number of training samples and 
time required for inference.
In this paper, we propose an end-to-end learning solution to generate 6-DOF parallel-jaw grasps starting from the 3D partial view of the object.
Our Learning to Grasp (L2G) method gathers information from the input point cloud through a new procedure that combines a differentiable sampling strategy to identify the visible contact points, with a feature encoder that leverages local and global cues.
Overall, L2G is guided by a multi-task objective that generates a diverse set of grasps by optimizing contact point sampling, grasp regression, and grasp classification.
With a thorough experimental analysis, we show the effectiveness of L2G as well as its robustness and generalization abilities.

\end{abstract}

\begin{figure*}[h!]
    \centering
    \includegraphics[width=0.85\textwidth]{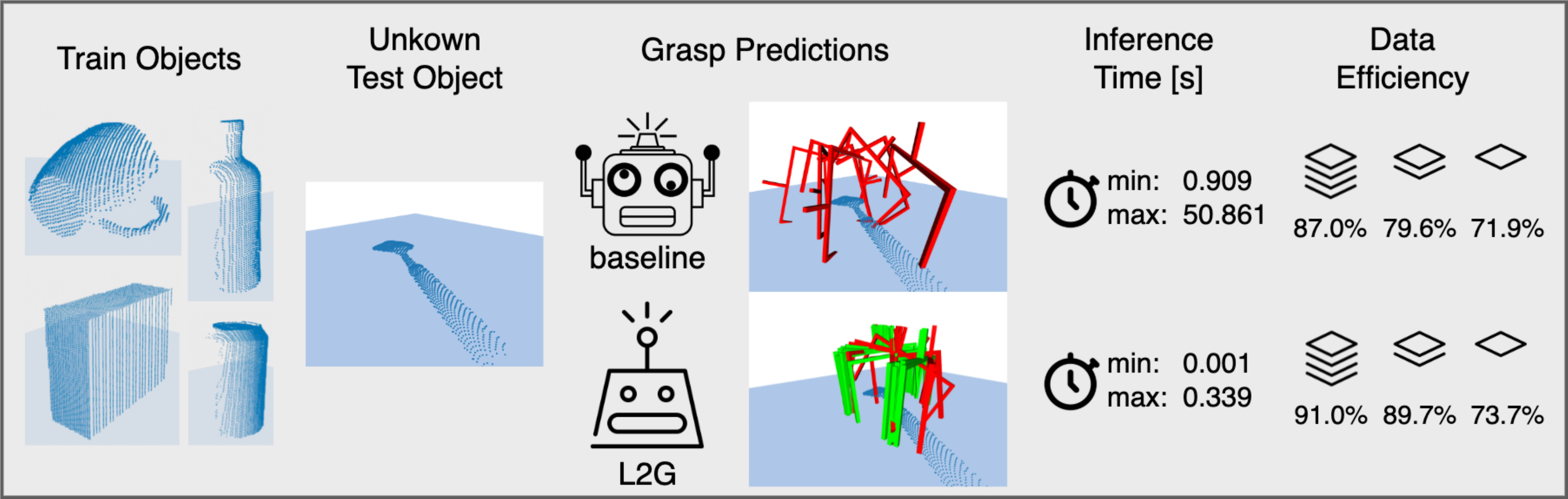}
    \caption{Overview of the advantages of our L2G model over the GPNet~\cite{WuChenNeurIPS20} baseline.
    Trained with the same training dataset, L2G is able to predict a diverse set of reliable grasps even for objects from unseen categories and different shape distributions.
    The inference times are much shorter and drastically facilitate real-world application.
    Finally, L2G is showing higher grasp success rates even when trained with only half the amount of data.
    }
    \label{fig:teaser}
    \vspace{-4mm}
\end{figure*}


\section{INTRODUCTION}
Grasping and manipulating unknown objects in unstructured, real-world environments is a long-standing challenge in robotics research.
Ideally, we would like robots to be able to observe 3D objects and propose a variety of reliable grasps, out of which collision-free and kinematically feasible actions can be executed. However, there are many challenges in the whole grasping pipeline that need to be tackled, from perception to planning and control. 
Some works have simplified the task by focusing on 2D perception and planar grasps prediction: a camera observes the scene perpendicularly and the gripper pose is constrained to be 
parallel to the image plane~\cite{dexnet2,morrison2018closing}. In this way, important geometric information may be disregarded inducing a limit in the grasp quality: the final effect is that the learned models hardly generalize beyond the scenario seen during training.
When dealing with 3D perception, a very first bottleneck is due to imprecision and deficiency in sensing: the information acquired from the observed scene is usually noisy and affected by variations in the environmental conditions.
Several works have proposed to overcome these issues by exploiting extra sources of geometric and physical information about the observed objects~\cite{modelbased1,modelbased2}, but these are not generally applicable for unknown shapes.
Other approaches rely on specific gripper information which makes them less ready to generalize in real-world applications \cite{contactgraspnet}.
Besides being very sensitive to shape variations (dimension and aspect ratio), 3D-based existing approaches have high sample and time complexity: they need a large number of densely annotated data to be trained \cite{graspness,regnet} and a long prediction time when deployed \cite{WuChenNeurIPS20}. This is mainly due to the use of handcrafted space quantization strategies and other heuristics that need to be progressively adjusted while training.
From the implementation point of view, the approaches that take point clouds as input are often cumbersome.
Although described as \textit{end-to-end} strategies, they are multi-stage techniques composed of networks trained in cascade or simultaneously with separate losses: each network has its own learning objective with no gradient flow among each other~\cite{grasp_pose_refinement,regnet}. 
Furthermore, these works mostly concentrate on feature extraction from local point neighborhoods, with little consideration given to the global appearance of the point-cloud.

With our work (see Fig.~\ref{fig:teaser}) we aim at pushing deep learning models for robot grasping one step further by overcoming at once the limitations described above. %
We introduce \textbf{\textit{Learning to Grasp (L2G)}}, an efficient end-to-end learning strategy to \tat{propose} 6-DOF parallel-jaw grasps starting from a partial point cloud of an object. Differently from previous work, our approach does not exploit any geometric assumption or impose gripper constraints.
\textbf{L2G builds on a differentiable sampling procedure. Specifically, it is guided by a multi-task optimization objective that identifies a set of diverse and reliable grasps by solving contact point sampling jointly with grasp regression and grasp classification.}
We show how L2G largely improves over its competitors with an advantage that becomes ever more evident when reducing the amount of available training data. It also demonstrates remarkable generalization ability in challenging settings where training and test data present significant shape variations as well differences in the gripper. \textbf{Moreover, we go beyond the use of standard backbone architectures discussing how a self-supervised pre-trained encoder that combines local and global cues can be easily plugged into the network and provides a further advantage.}

\section{RELATED WORK}
The task of grasping rigid objects with a 2-finger gripper consists in identifying the pose of the gripper in which the fingers close, starting from some representation of the object. 

\textit{Data and representations.}
A large part of the grasping literature has focused on 2D and 2.5D (images with depth maps) data \cite{morrison2018closing,roibased,fcrgbd}. This setting simplifies the definition of the grasping problem, but at the same time limits its applicability with gripper forced to approach objects vertically. Dealing with images provides the possibility to exploit supervised Imagenet pre-trained networks as well as large scale self-supervised models reducing the need for data annotation \cite{zhu2022grasp, huang2020dipn, zeng2018learning}. On the other hand,
point clouds allow better reasoning on the geometric properties of the objects and more freedom for the gripper pose \cite{WuChenNeurIPS20,qin2020s4g, pointnet2grasp,mousavian2019graspnet, murali2020}, but 3D grasping is more challenging and needs densely annotated data.
Here transfer learning from supervised or self-supervised pre-trained models \cite{Alliegro_2021_CVPR,OcCo} is not standard practice, and existing works
based on point clouds mainly exploit PointNet \cite{pointnet_CVPR17} and PointNet++ \cite{NIPS2017_pointnet2} representations which focus on local information, lacking the ability to properly capture the global object shape.

Several works have been dedicated to collecting and annotating grasping datasets by exploiting physical grasps in simulation engines \cite{database,databaseJacquard,WuChenNeurIPS20}. Most recent publications have also proposed larger testbeds, but their simulation environments have not been released yet, which makes it difficult to consider the methods proposed in the same papers as benchmark reference \cite{mousavian2019graspnet,liang2019pointnetgpd}, \cite{contactgraspnet}.

\textit{Grasping Methods.} 
The earlier grasping approaches were based on handcrafted features \cite{handcrafted1,handcrafted2}, while in recent years data-driven methods have gained popularity \cite{Kleeberger2020}. They can be categorized as model-based and model-free: the former relies on object-specific knowledge such as a 3D model or surface characteristics \cite{modelbased1,modelbased2,dexnet1}, while the latter assumes that no such explicit information is available. Model-free methods infer grasp poses purely based on the perceived information, and they can be conveniently applied to novel objects for which specific models are not available.  Among them, Deep-Learning-based discriminative strategies evaluate a given set of grasp candidates  \cite{dexnet2,liang2019pointnetgpd}.
On the other hand, generative approaches regress the best grasp poses, however they usually lack grasp diversity \cite{generative1,generative2}.
The most recent {grasping methods} combine the two aspects and incorporate both generative and discriminative components.
In particular, \cite{mousavian2019graspnet, murali2020} generate grasp poses by means of a variational autoencoder trained on the distribution of successful grasps obtained from a simulation engine. A subsequent classifier and an iterative grasp refinement are used to rank and improve the candidates.
In \cite{qin2020s4g, pointnet2grasp}, the authors use a tailored grasp representation and regress a grasp with a relative graspability score for each point of the input point cloud.
A coarse grasp for every point is also predicted and then refined in \cite{grasp_pose_refinement}. A combination of sub-networks for point selection and refinement is used in \cite{regnet}. The recent work \cite{graspness} relies on dense grasp annotations and predicts the probability of successful grasping for each point which are then subselected via Farthest Point Sampling (FPS), while in \cite{contactgraspnet} FPS is applied upfront to limit memory cost before running grasp prediction for each obtained point.
GPNet \cite{WuChenNeurIPS20} uses a grid-based heuristics to generate grasp proposals. 
A discrete set of regular 3D grid points is defined and proposals are obtained by pairwise combination of all input points (as contact points) with all grid points (as grasp centers).
The learning process exploits an antipodal classifier to reduce the large number of proposals, a regression module to predict approach angles and offsets to the grasp centers, and a grasp classification module to estimate the success likelihood of the regressed grasps. Finally, the large number of predictions is reduced by Non Maximum Suppression (NMS).

Our L2G method fits in the context of Deep-Learning-based models for grasping from object point clouds. 
Given a partial observation of an unknown object, we \textit{learn to sample} a set of suitable contact points and proceed to predict a 6-DOF grasp only for these points.
Our strategy avoids unnecessary overhead by sub-optimal heuristics, na\"ive FPS, or expensive refinement stages and post-hoc NMS. Moreover, we show how 3D self-supervision can be leveraged to improve data representation with a beneficial effect on grasping performance.

\section{METHOD}
\begin{figure}[tb]
    \centering
    \includegraphics[width=0.48\textwidth]{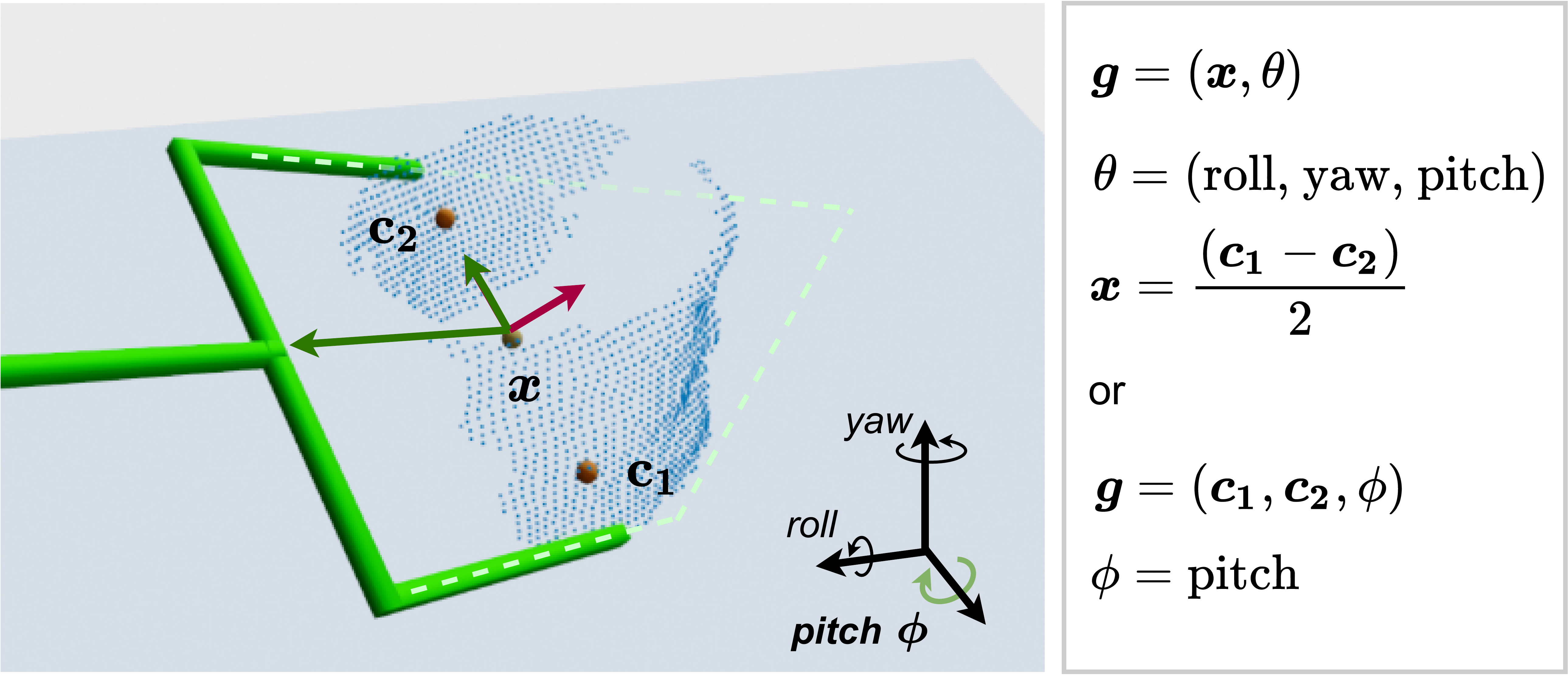}
    \caption{The world coordinate system has its horizontal plane parallel to the ground and the vertical axis orthogonal to it. The grasp configuration is defined by the two contact points $(\bc_1,\bc_2)$ and the pitch angle $\phi$ which is the angle formed by the plane of the gripper (dashed lines) with the horizontal plane. We indicate with $\bx$ the center between the contact points, while $\theta \in [-\pi,\pi]^3$ combines the information of yaw, roll, and pitch.} \vspace{-4mm}
    \label{fig:coordinates} 
\end{figure}

\begin{figure*}[tb]
    \centering
    \includegraphics[width=0.9\textwidth]{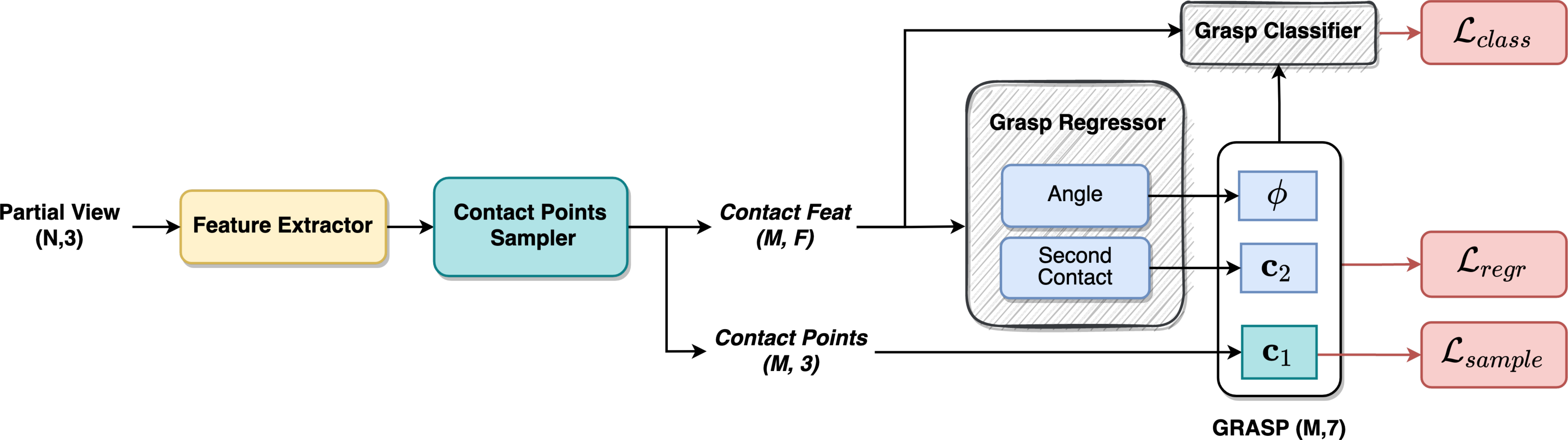}\vspace{-2mm}
    \caption{Schematic overview of our Learning to Grasp (L2G) multi-task network. The \textit{feature extractor} corresponds to our backbone encoder: we use both PointNet++ \cite{NIPS2017_pointnet2} and \deco \cite{Alliegro_2021_CVPR}.
    The \textit{contact point sampler} produces points close to the ground truth contact points while soft-projecting them on the object surface. 
    The \textit{grasp regressor} and the \textit{grasp classifier} are inherited from \cite{WuChenNeurIPS20}: the former  predicts the second contact point and the pitch angle for each grasp, while the latter scores the predicted grasps to identify the most reliable ones.
    }
    \vspace{-4mm}
    \label{fig:architecture} 
\end{figure*}

The core contribution of our L2G model is in the procedure used to gather information from the input point cloud which combines the sampling procedure adopted to identify the contact points, and the DeCo feature encoder \cite{Alliegro_2021_CVPR}.
Overall L2G is formalized as a multi-task deep architecture as shown in Fig. \ref{fig:architecture}.

\subsection{Problem Statement}

Given a point cloud $\sP=\{\bp_i \in \R^3\}_{i=1}^N$ representing the visible surface of an object, we indicate a parallel-jaw grasp as $\bg=\{\bx,\theta\}\in SE(3)$. Here $\bx \in \R^3$ locates the center of the two parallel jaws, and $\theta\in[-\pi,\pi]^3$ is the Euler angle describing the 3D orientation of the gripper, which can be also identified by its unit quaternion representation $\bu=quat(\theta)$.
A grasp can be alternatively defined by $\bg=\{\bc_1, \bc_2, \phi\}\in \R^7$. Where $(\bc_1,\bc_2)$ are the two contact points on the object surface, which determine the grasp center $\bx$ as well as the roll and yaw orientation of $\theta$, while $\phi \in [0,\pi]$ is the remaining pitch orientation corresponding to the gripper approach angle\footnote{Note that by restricting $\phi$ to $[0,\pi]$ instead of $[-\pi,\pi]$, we eliminate grasps with an approach vector in the lower hemisphere for which the gripper would collide with the ground plane.}
 (see Fig.~\ref{fig:coordinates}).
A physical simulation engine provides us with a set of positive $\sG^+$ (label $l=1$) and negative  $\sG^-$ (label $l=0$) ground truth grasps for $\sP$. They all satisfy the antipodal constraint~\cite{Nguyen1988}, but the negative ones fail to successfully lift the object. This may be due to collisions of the gripper with the object or ground, not making proper contact, or object slipping during lifting.
We formalize the task of visual \textit{learning to grasp} as learning the mapping from the object point cloud $\sP$ to the set of ${\bg}_{j=1}^M \in {\sG}$ grasps that best matches $\sG^+$.
In order to avoid ambiguities due to symmetry, during evaluation all grasps are mapped into an unambiguous half-space by considering the contact point closer to the ground plane as $\bc_1$, as done in \cite{WuChenNeurIPS20}.

\subsection{Learning to Grasp}
Our L2G architecture is optimized to predict all the components (contact points and pitch angle) of a grasp at once.
It is composed of Feature Extractor, Contact Point Sampler, Grasp Regressor, and Grasp Classifier. 
Each of the modules is described in detail in the next paragraphs.

\vspace{0.5mm}\textit{Feature Extractor $(\R^{N \times 3}\rightarrow \R^{N \times F})$}. 
The feature extractor learns an $F$-dimensional representation for each point of the observed point cloud $\sP$.  We consider two possible encoders: a standard PointNet++ \cite{NIPS2017_pointnet2} and \deco~\cite{Alliegro_2021_CVPR}. The latter exploits graph convolutions and combines local information from denoising with global information from contrastive learning: the encoder is pre-trained with these two self-supervised tasks and has shown remarkable results when used for shape completion.

\vspace{0.5mm}\textit{Contact Point Sampler $(\R^{N \times F}\rightarrow \R^{M \times F})$}. The goal of the sampler is to identify the set of reliable contact points $\sQ=\{\bq_j \in \R^3\}_{j=1}^M$ with $M\leq N$, out of the visible point cloud $\sP$, and collect for each of them the corresponding $F$-dimensional representation vector. The major issue with sampling is that it is a non-differentiable operation, but recent papers have proposed effective workarounds \cite{Dovrat_2019_CVPR,lang2020samplenet}. We leverage these approaches to produce an $M\times3$ matrix which can be interpreted as a set of $M$ points. Specifically, we learn to produce $M$ points that are close to the set of visible contact points $\sC$ of $\sG^+$ such that, for each of them, the projection on the object surface soft-matches to a single point of $\sP$.
The first goal is attained by optimizing the average nearest neighbor loss 
\begin{equation}
    \sL_{nn}(X,Y) = \frac{1}{|X|}\sum_{\bx\in X} \min_{\by \in Y} \|\bx - \by\|^2_2~,
\end{equation}
and the maximal nearest neighbor loss
\begin{equation}
    \sL_{mn}(X,Y) = \max_{\bx \in X}\min_{\by \in Y} \|\bx - \by\|^2_2~,
\end{equation}
combined in the following closeness-coverage loss
\begin{equation}
    \sL_{cc}(\sQ,\sC) = \sL_{nn}(\sQ,\sC)+\sL_{nn}(\sC,\sQ) + \sL_{mn}(\sQ,\sC)~.   
\end{equation}
Here the first and last term forces the produced points to stay \textit{close} to the grasp contact points both in average and in the worst case, while the second term ensures the full \textit{coverage} of the grasping input set. 
Furthermore, for each point $\bq$ we search the set $\bp_{i=1}^{k} \in \sN_{\sP}(\bq)$ of its nearest neighbors from $\sP$ in terms of Euclidean distance $d_i=\|\bq-\bp_{i}\|_2$. The $k$ neighbors are used to evaluate the projection $\br$ of the produced point $\bq$ on the object surface, formalized by the following linear combination 
\begin{equation}
    \br = \sum_{\bp_i\in\sN_{\sP}(\bq)}\omega_i\bp_{i}~,
\end{equation}
where 
\begin{equation}
    \omega_i=\frac{\exp^{-d_i^2/t^2}}{\sum_{\bp_i\in\sN_{\sP}(\bq)}\exp^{-d_i^2/t^2}}~.
\end{equation}
These weights can be intended as a probability distribution over the points in $\sP$, guided by the temperature parameter $t$. For high temperature values, the distribution becomes more and more uniform, while for low temperature values the distribution collapses to a Kronecker delta on the closest point. This last condition mimics the desired sampling and can be obtained by minimizing the projection loss:
\begin{equation}
\sL_{proj}=t^2~.
\label{eq:projloss}
\end{equation}
Finally the sampling loss is $\sL_{sample} = \alpha \sL_{cc} + \sL_{proj}$~.

\vspace{0.5mm}\textit{Grasp Regressor} $(\R^{M \times F}\rightarrow \R^{M \times 4})$. Starting from the features of each selected point, we rely on the simplified hypothesis that it corresponds to only one possible successful grasp, and the grasp regression module predicts both its second contact point ($\bc_2 \in \R^3$) and the grasp pitch angle ($\phi \in \R^1$).
The learning process is guided by a loss that measures the distance between each predicted grasp ${\bg}_j=({\bc}_1, {\bc}_2,{\phi})_j$ and its closest ground truth grasp $\bg^+_j=(\bc_1^+, \bc_2^+,\phi^+)_j$. 
Specifically, the ground truth points~$\bc_1^+$ are sorted based on their distance to ${\bc}_1$, and the closest one identifies the reference ground truth grasp. 
The distance between the grasps is measured in terms of the position of their centers and variation of the corresponding angles defined in terms of the quaternion representation~$\bu$:
\begin{equation}
    \sL_{regr} = \frac{1}{M}\sum_{j=1}^M \|{\bx}_j - \bx^+_j\|_2 +  \lambda\arccos(|\langle {\bu}_j,\bu_j^+\rangle|)~,
\end{equation}
where $\lambda$ weighs the contributions of Euclidean and angular distances, as similarly done in \cite{EppnerISRR2019}.

\vspace{0.5mm}\textit{Grasp Classifier $(\R^{M \times 4} + \R^{M\times F}  \rightarrow \R^{1})$}. The grasp classifier takes as input the information on the second contact point and angle ($\R^{4}$) as well as the features of the first contact point ($\R^F$) to finally score the grasp. 
Its purpose is to sort the predicted grasps and eliminate those that are unlikely to succeed.
We use a simple binary cross-entropy loss where we indicate the predicted output with ${s}_j\in \R^1$ and the ground truth label with $l_j=[0,1]$:
\begin{equation}
    \sL_{class}=-\frac{1}{M} \sum_{j=1}^M (l_j \log{{s}_j} + (1-l_j)\log(1-{s}_j))~.
\end{equation}

All the loss contributions guide jointly the training process of our L2G: $\sL= \sL_{sample} + \sL_{regr} + \sL_{class}$.

\subsection{Implementation Details}
In the previous section, we provided a high-level intuition about the internal functioning of our approach by referring to a generic $F$-dimensional feature vector. Here we describe the architecture, the learned intermediate embeddings, and the hyperparameters of our model in more detail.

Our L2G employs the same feature extractor as GPNet \cite{WuChenNeurIPS20}: a PointNet++ with four multiscale-grouping Set Abstraction~(SA) layers followed by four Feature Propagation~(FP) layers. For each observed partial object point cloud we obtain the global feature vector $F_s \in \R^{1024}$ by performing max-pooling on the feature map output of $\textit{SA}_{4}$. The per-point features $F_p \in \R^{128}$ are obtained right after $\textit{FP}_4$.

We dub our model \textit{L2G+DeCo} when using as feature extractor the two-branch graph-convolutional backbone presented in \cite{Alliegro_2021_CVPR}, pre-trained via self-supervision on  ShapeNetPart~\cite{Yi_ACM_2016_shapenetpart}. In this case, the global feature vector $F_s \in \R^{1024}$ is obtained from the global encoder branch, while the local encoder output is first concatenated with the global vector and then processed through a 128-dimensional convolutional layer. The output defines the per-point features $F_p \in \R^{128}$.

The contact point sampler takes as input the feature vector $F_s$ to produce (soft sampling at training time) the $\bq_{j=1}^M$ points of the $\sQ$ set. The sampler is composed by four Fully Connected layers followed by the soft-projection module which has the same structure described in \cite{lang2020samplenet}.
We consider $M=500$ and use a small local context for the projection loss by setting the size of neighborhood $\sN_{\sP}(\bq)$ to $k=10$.
For each  $\bq$, we further define $\sN_{\sF}(\bq)$ by grouping the per-point feature vectors $F_p$ of the $nn=100$ nearest points on the input shape.
The grasp regressor is fed with input $\sN_{\sF}(\bq)$ and predicts both the second contact point and the angle ($\bc_2, \phi$).
The grasp classifier combines  $\sN_{\sF}(\bq)$ with $(\bc_2,\phi)$.
Specifically, an MLP layer takes as input $(\bc_2,\phi)\in \R^4$ to get a 128-dimensional feature vector which is aggregated by summation with $\sN_{\sF}(\bq)$ before entering a second MLP layer with 1-dimensional output followed by a sigmoid function.
For a detailed analysis of the hyperparameters $M$ and $nn$ we refer to the next section and specifically Fig.~\ref{fig:sampling-horizontal}.

In all our experiments we set the loss weighting parameters to $\alpha=10$ and $\lambda=0.1$.
For each experiment, we take the average of three runs with different seeds and report the results from the last epoch.
Our code is implemented in PyTorch~1.8 with CUDA~11.1. Our models are trained on a single NVIDIA Tesla V100 16GB GPU, inference and real-world experiments are performed on an NVIDIA 2080 GPU. 
The code, pre-trained models, and data are available at \url{https://github.com/antoalli/L2G}. The project page also presents video examples of our robotic experiments.

\begin{table*}[t]
\caption{Left: Simulation-based and Rule-based evaluation results on ShapeNetSem-8. GPNet refers to the top results in \cite{WuChenNeurIPS20}. GPNet* indicates the results obtained by re-running the authors' code. Right: Simulation-based success rate in relation to the coverage for different sensitivity settings.
}
\begin{minipage}[tb]{0.68\linewidth}
\label{tab:sota} \vspace{-4cm}
\resizebox{1\textwidth}{!}{
\begin{tabular}{c | c|c|c|c || c|c|c|c | c|c|c|c}
\hline
\multicolumn{5}{c||}{Test Set from \cite{WuChenNeurIPS20} - Simulation Based} & \multicolumn{8}{c}{Test Set from \cite{WuChenNeurIPS20} - Rule Based} \\
\hline
\multirow{2}{*}{Method} & \multicolumn{4}{c||}{success rate @k\%} & \multicolumn{4}{c|}{success rate @k\%} & \multicolumn{4}{c}{coverage rate @k\%} \\
\cline{2-13}
& 10 & 30 & 50 & 100 & 10 & 30 & 50 & 100 & 10 & 30 & 50 & 100\\
\hline
GraspNet \cite{mousavian2019graspnet} & 80.0 & 59.4 & 50.8 & 35.4 & 86.7 & 83.3 & 73.3 & 53.4 & 6.3 & 6.3 & 12.2 & 16.8\\
GPNet \cite{WuChenNeurIPS20}  & 90.0 & 76.1 & 72.3 & 58.8 & 93.3 & 93.2 & 82.0 & 72.9 & 6.8 & 14.4 & 19.9 & 30.7\\
GPNet* & 92.2 &	90.0 &	82.3 &	59.7 & 91.1 &	89.5 &	85.0 &	67.5 &	7.6 & 	15.2 &	25.3 &	34.5\\
L2G & 93.6	& 90.1	& 87.9	& 82.0 & 94.8 &   \textbf{95.9} &	\textbf{95.3} &	\textbf{95.1} &	19.2 &	29.1 &	34.5 &	39.9\\
\hline
GPNet*+DeCo \cite{Alliegro_2021_CVPR} & 91.1 &	89.1 &	81.9 & 67.0 & 89.4 &	85.0 &	82.5 &	70.4 &		7.2 &	15.1 &	24.6 &	33.9 \\
L2G+DeCo \cite{Alliegro_2021_CVPR} & \textbf{94.6} & \textbf{93.5} &	\textbf{91.4} &	\textbf{82.9} & \textbf{95.2} &	94.9 &	94.6 &	94.5 &	\textbf{20.6} &	\textbf{29.2} &	\textbf{35.5} &	\textbf{41.8} \\
\hline \hline
\multicolumn{5}{c||}{Extended Test Set - Simulation Based} & \multicolumn{8}{c}{Extended Test Set - Rule Based}\\ 
\hline
GPNet* & 87.0 &	85.4 &	79.2 &	59.5 & 89.6	& 87.9	& 83.7	& 68.1	&	7.6&	15.5 &	24.5 &	34.4\\
L2G & 91.0 & 89.0 &	87.2 &	81.4 & 94.9	& 96.0	& 96.2	& \textbf{95.8}	&	19.1 &	29.1 &	34.0 &	39.8 \\
\hline
GPNet*+DeCo \cite{Alliegro_2021_CVPR}  & 91.2 &	89.5 & 83.1 &	68.5 & 89.2 &	86.3 &	82.9 &	72.3 &	7.5 &	15.9 &	25.0 &	34.7  \\
L2G+DeCo \cite{Alliegro_2021_CVPR} & \textbf{93.0} & \textbf{92.1} & \textbf{90.7} & \textbf{83.2} & \textbf{96.4} & \textbf{96.5} &	\textbf{96.4} &	95.4 &	\textbf{20.3} &	\textbf{31.0} &	\textbf{36.5} &	\textbf{41.5}\\
\hline
\end{tabular}
}
\end{minipage}\hspace{5mm}
 \includegraphics[width=0.3
    \textwidth]{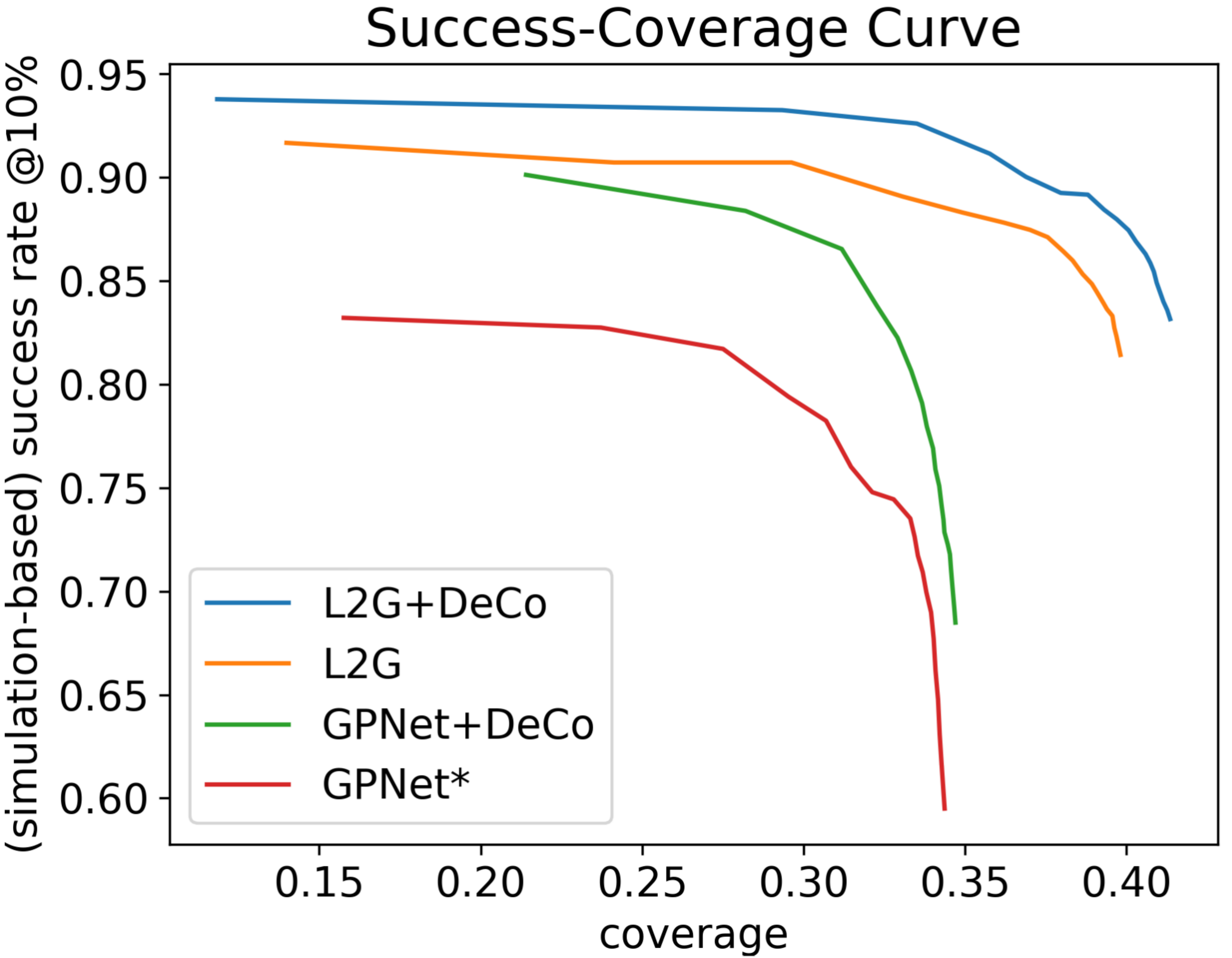} 
    \vspace{-5mm}
\end{table*}

\section{EXPERIMENTS}
\subsection{Datasets, Metrics and Baselines}
For our experiments, we consider two datasets.
\textit{ShapeNetSem-8}  \cite{WuChenNeurIPS20} consists of 226 CAD models of 8 object categories (bowl, bottle, mug, cylinder, cuboid, tissue box, soda can and toy car) from ShapeNetSem~\cite{ShapeNetSem}. Each object comes with $\sim$100k annotated grasps and associated grasp success or failure label obtained with the Pybullet physics simulator~\cite{pybullet}. 
The dataset is split into 196 object instances for training and 30 object instances for testing.
The test set in~\cite{WuChenNeurIPS20} was composed of only one view per object.
To get more statistically meaningful results we also present an \textit{Extended Test Set} of 5~different views per object, with each view serving as an independent test sample.

Using the Pybullet simulator, we also created our second dataset with 76 objects from YCB~\cite{YCB}, dubbing it \textit{YCB-76}. Each object is placed in various stable resting poses, totaling 259 distinct grasping scenarios, with point clouds generated from 10 arbitrary views. 
We consider the described YCB-76 as test set after having trained the grasping models on ShapeNetSem-8. For most of the objects, we only focused on the grasping success in simulation, but for a subset of them (\textit{YCB-8}: cracker box, mug, tomato soup can, potted meat can, mustard bottle, flat screwdriver, large clamp, tennis ball) we also collected $\sim$100k ground truth grasps in the same fashion as in ShapeNetSem-8 to run a detailed rule-based analysis.

We perform our experimental evaluation by using the same metrics of \cite{WuChenNeurIPS20}: simulation-based success rate as well as rule-based success rate and coverage.
For the former, we use the same simulation environment used for grasp annotation.
The rule-based metrics instead compare a predicted grasp $\bg$ to the reference annotated grasps $\bg^+ \in \sG^+$ by means of Euclidean $\delta_{\bx}(\bg, \bg^+) = || \bx - \bx^+ ||_2$ and angular $\delta_{\theta}(\bg, \bg^+) = \arccos{|\langle \bu,\bu^+ \rangle|}$ distances.
A prediction is considered successful if there exists at least one $\bg^+$ with $\delta_{\bx}\leq 25mm$ and $\delta_{\theta}\leq 30^{\circ}$. 
Conversely, a grasp annotation $\bg^+$ is considered covered, if there is a prediction ${\bg} \in {\sG}$ close by, using the same distance criterion.
Hence, the coverage expresses what fraction of $\sG^+$ is covered by the grasp predictions in ${\sG}$.
{The rule-based success rate may be overly optimistic because it only takes into account proximity to successful ground truth grasps but not proximity to unsuccessful ones.} 
For all three metrics, we consider the predictions ranked in the top $\text{k} = \{10, 30, 50, 100\}\%$, reporting the values as @k\%.

As reference baselines we consider GPNet \cite{WuChenNeurIPS20} and GraspNet \cite{mousavian2019graspnet}. In particular the former is our best competitor: we highlight that this approach, besides relying on a heavy initial space quantization to choose the reference contact points, also requires an 
NMS post-processing stage to refine the predicted grasps.
Moreover, to guarantee a fair comparison, we report both the results of GPNet from the original paper and our re-run of the authors' code indicated as GPNet*. With \textit{GPNet+DeCo} we refer to the baseline where we applied the encoder from~\cite{Alliegro_2021_CVPR}.

\subsection{Experiments on ShapeNetSem-8}
We start our analysis by running experiments on the training and test data of ShapeNetSem-8. 
The results reported in Table~\ref{tab:sota} (left) show that L2G and L2G+DeCo consistently outperform the corresponding GPNet and GPNet+DeCo baselines. 
On the smaller test set of \cite{WuChenNeurIPS20}, the effect of the DeCo encoder on top of L2G might be marginal. 
Instead, the advantage of DeCo becomes clear for both GPNet and L2G  when focusing on  the simulation-based results of the extended test set which should be considered the more reliable testbed.
Notably, by increasing @k\%, \ie when taking into account predictions with lower confidence, the performances of the baselines drop significantly, whereas L2G maintains high success rates.
By combining this information with the increased coverage rate we can conclude that L2G is effectively providing a wide variety of reliable grasp predictions and that DeCo further enhances this effect.
The success-coverage plot in the right part of Table~\ref{tab:sota} confirms this behavior. It is comparable to a precision-recall curve where the simulation success rate resembles the precision and the coverage is the recall. 

\begin{figure*}[tb]
    \centering \vspace{-2mm}
    \begin{tabular}{@{}cc@{}c@{~}c@{}}
        \hspace{-2mm}\includegraphics[height=0.18\textwidth]{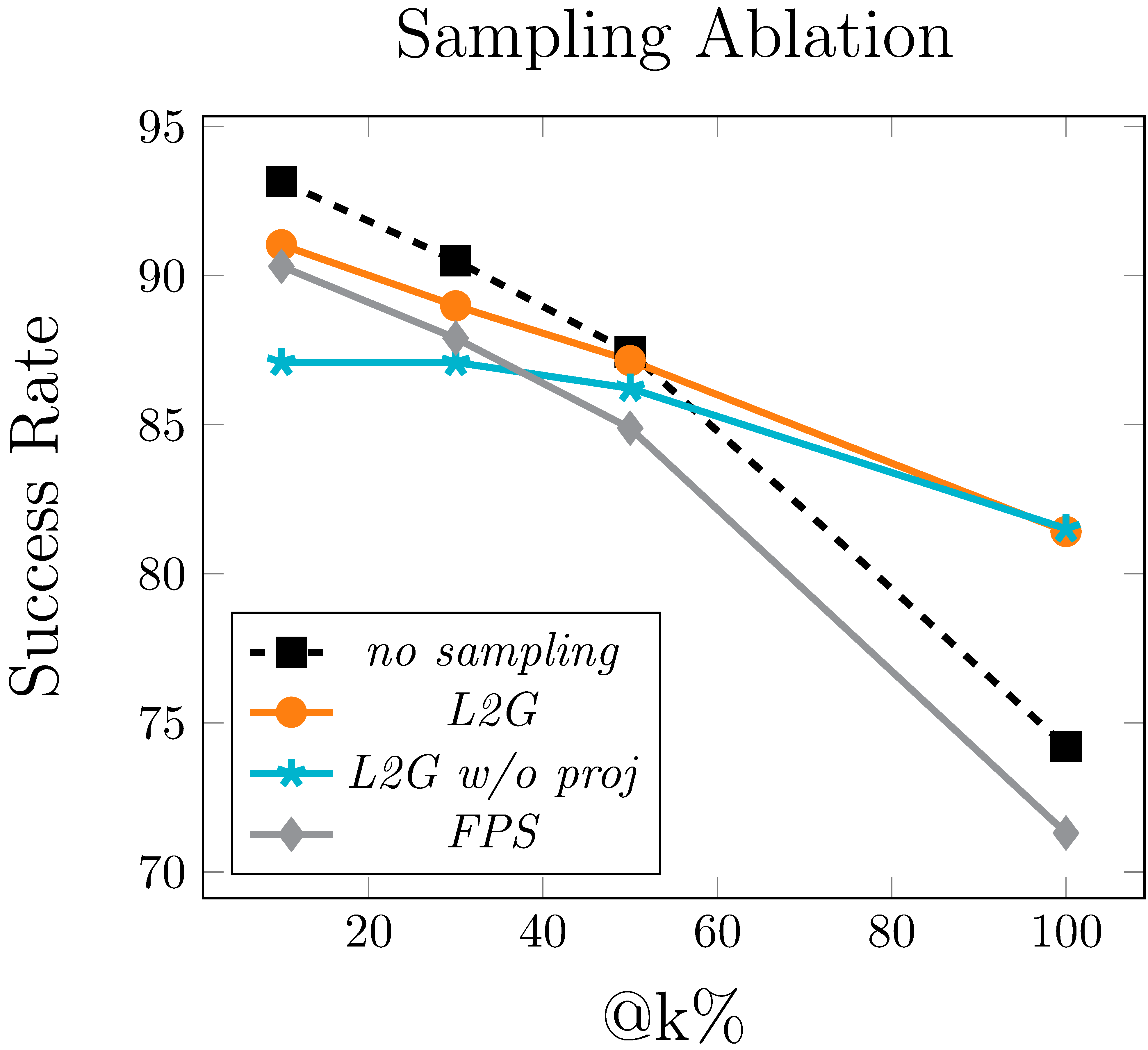} &
        \includegraphics[height=0.18\textwidth]{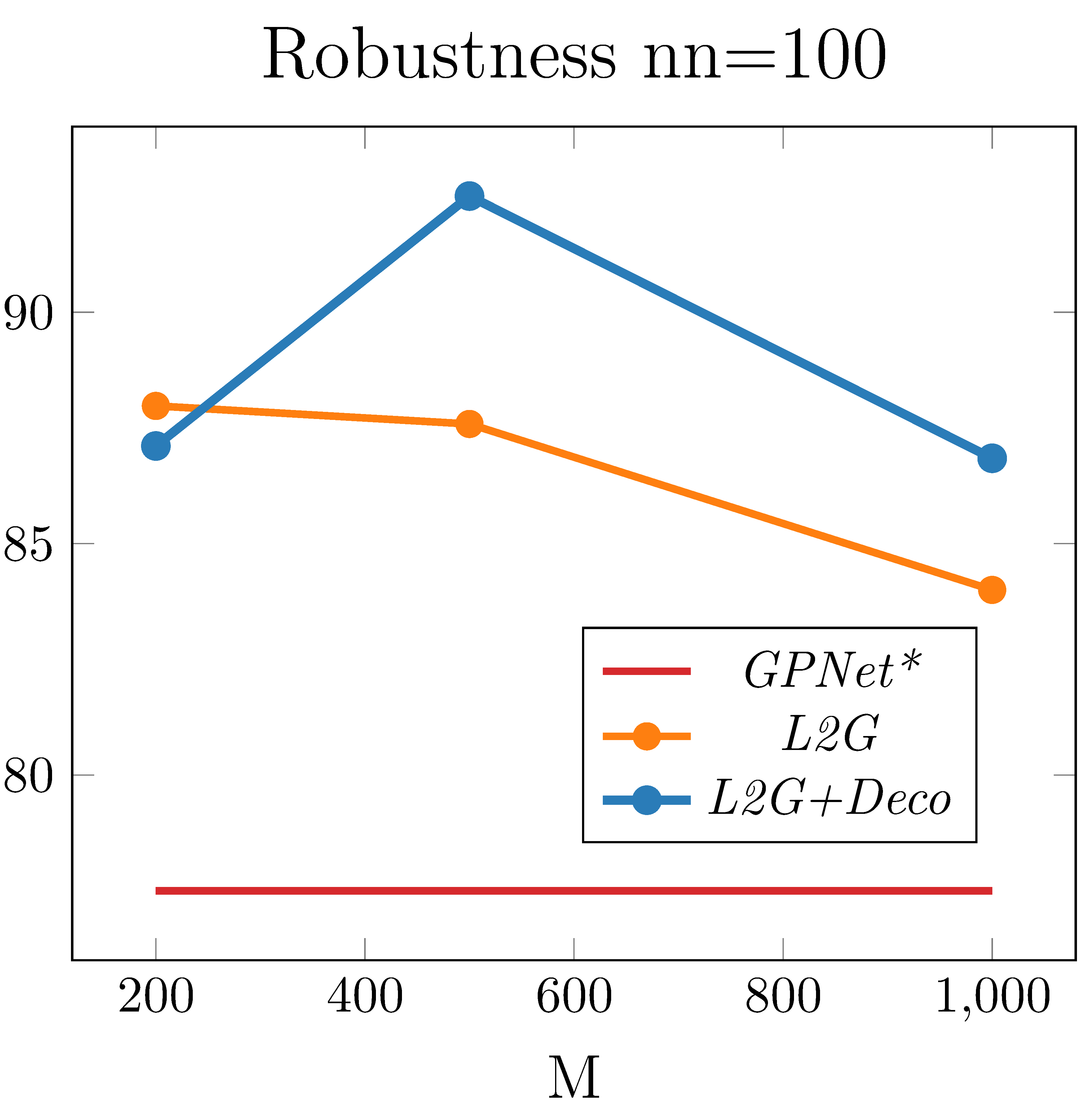} & 
        \includegraphics[height=0.18\textwidth]{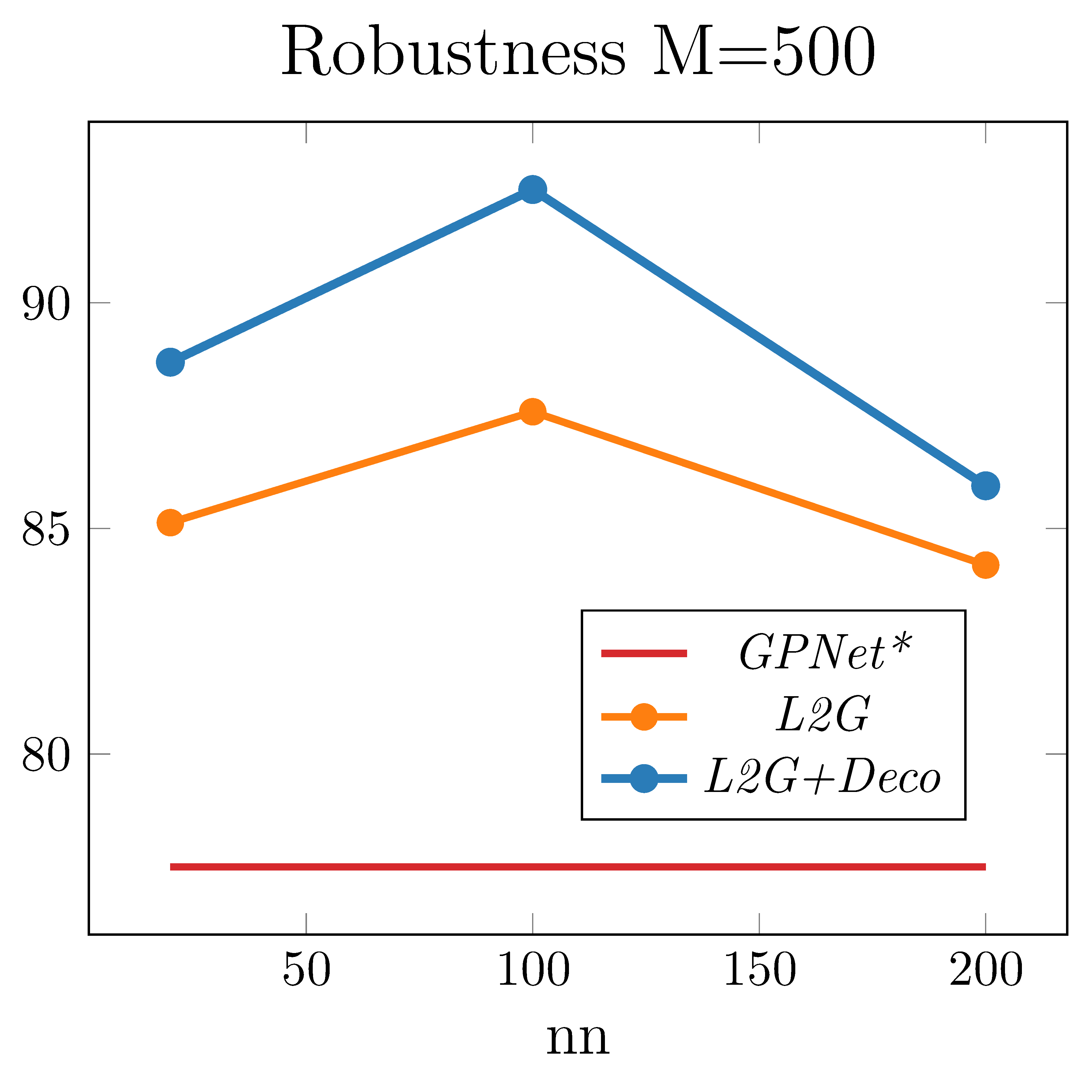}&
        \includegraphics[height=0.18\textwidth]{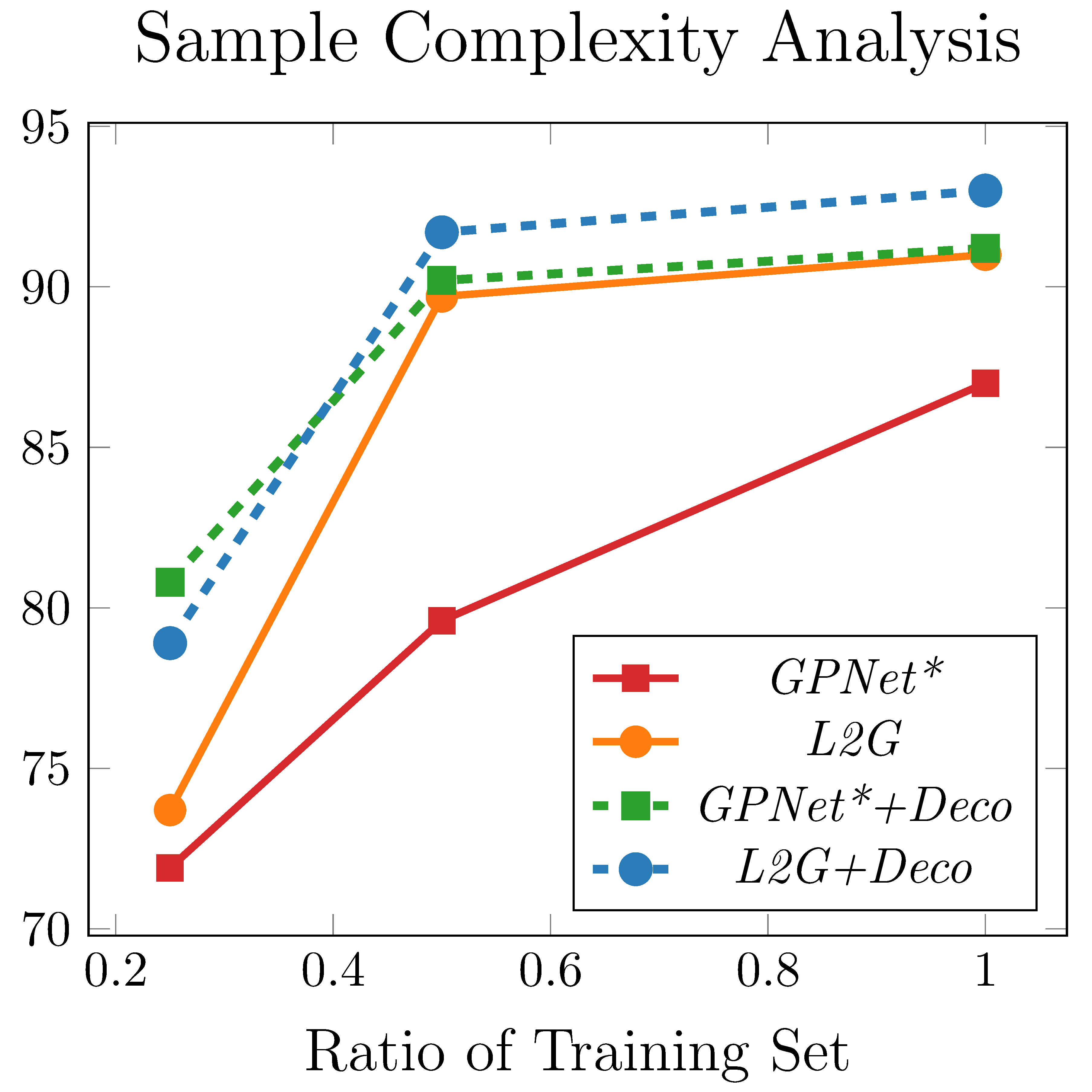} 
    \end{tabular}
    \resizebox{0.21\textwidth}{!}{
    \begin{tabular}{@{~}cc@{~}c@{~}}
    \hline
    \multicolumn{3}{c}{Time Complexity Analysis}\\
    \hline
    \multirow{3}{*}{Method} & \multicolumn{2}{|c}{Inference Time } \\
    & \multicolumn{2}{|c}{per shape (s)} \\
    \cline{2-3}
    & \multicolumn{1}{|c}{min} & max \\
    \hline
    \multicolumn{1}{c|}{GPNet*}  & 0.909 & 50.861\\
    \multicolumn{1}{c|}{L2G} & \textbf{0.001} & \textbf{0.339}\\
    \hline
    \multicolumn{1}{c|}{GPNet*+DeCo} & 0.935 & 46.897\\
    \multicolumn{1}{c|}{L2G+DeCo} & 0.016 & 0.365\\
    \hline
    \end{tabular}
    }
    \caption{All the results refer to the simulation-based experiments on the extended test set of ShapeNetSem-8. First plot: ablation analysis to study the role of the contact point sampler. Second and third plot: robustness evaluation when varying $M$ and $nn$. Fourth plot: sample complexity analysis executed by observing the performance of the grasping methods when changing the amount of training samples. Table: time complexity analysis.}
    \label{fig:sampling-horizontal}
    \vspace{-4mm}
\end{figure*}

\subsection{Sampling Ablation}
We run an ablation analysis to assess the role of our contact point sampler.
Specifically, we compare L2G to a model that uses the same encoder and grasp heads but lacks the contact points sampler, resulting in a grasp prediction for each point in the observed point cloud (\textit{no sample}).
We also consider replacing our sampler with a not-learned \textit{FPS} selection process, as well as testing a variant of our L2G obtained by turning off $\sL_{proj}$ (\textit{L2G w/o proj}).
As shown by the first plot of Fig. \ref{fig:sampling-horizontal}, L2G w/o proj performs worse than L2G when considering the top 10, 30 and 50\% grasps, while on the entire set (100\%) the results are the same.
The success rate of FPS is always lower than L2G: such a na\"ive sampling process is the core of several 3D grasping approaches~\cite{contactgraspnet,regnet} and it is significantly less effective than our learned sampling.
Finally, the no-sampling strategy is the most expensive: we report it as a reference to show how our efficient L2G approaches its results and becomes equal or better when considering the top 50\% predicted grasps and the entire set.

\subsection{Robustness, Sample and Time Complexity Analysis}
We evaluate the robustness of L2G to its two hyperparameters:
the number~$M$ of sampled contact points which also corresponds to the total number of considered grasps, and the cardinality~$nn$ of the neighborhood $\sN_{\sF}(\bq)$ centered at each sampled contact $\bq$.
The second and third plot of Fig.~\ref{fig:sampling-horizontal} show that the performance of L2G has a mild dependence on $M$, with a decrease in success rate for very high values, while it is more robust to the choice of $nn$. The trend is similar also when using the DeCo encoder.
Both L2G and L2G+DeCo maintain their advantages over the GPNet baseline.

To investigate the sample complexity, we reduced the amount of training data by a factor  $0.25$ and $0.50$. 
The results in the fourth plot of Fig.~\ref{fig:sampling-horizontal} show that L2G outperforms GPNet even in low-shot scenarios. By comparing the accuracy of L2G at 0.25 (89.7) with that of GPNet at 1 (87.0) the low sample complexity of L2G appears even more evident. Similar conclusions can be drawn also comparing L2G+DeCo at 0.5 (91.7) with GPNet+DeCo at 1 (91.2).
DeCo well combines with the handcrafted space quantization strategy of GPNet at 0.25  producing the best results.

Finally, the table within Fig.~\ref{fig:sampling-horizontal} allows to compare GPNet and L2G in terms of minimum and maximum inference time, showing the significant advantage of our approach.

\subsection{Generalization on YCB}

To assess the generalization abilities of L2G we employ the YCB dataset as test set.
It has more object categories than ShapeNetSem-8 used for training.
Additionally, objects in YCB vary in dimension and appear in different resting poses, leading to grasping scenarios with a wide range of difficulty levels.
This is a challenging setting due to the need for overcoming both the semantic and the appearance domain shift.
The simulation-based and rule-based results on YCB-8 are in the left part of Table~\ref{tab3:ycb}, while the right part presents the simulation-based results on YCB-76.

\begin{table*}[t]
    \centering
    \caption{Results obtained when testing on YCB-8 and YCB-76. For the first, we collected grasp annotations to also assess the rule-based performance, while for the second we show only the simulation-based results.}
    \label{tab3:ycb}
    
    \resizebox{\textwidth}{!}{
    \begin{tabular}{c | c|c|c|c | c|c|c|c| c|c|c|c || c|c|c|c }
        \hline
        \multicolumn{13}{c||}{YCB-8} &   \multicolumn{4}{c}{YCB-76}
        \\
        \hline
        \multirow{3}{*}{Method} & \multicolumn{4}{c|}{Simulation Based} & \multicolumn{8}{c||}{Rule Based} & \multicolumn{4}{c}{Simulation Based} \\
        \cline{2-17}
        & \multicolumn{4}{c|}{success rate @k\%} & \multicolumn{4}{c|}{success rate @k\%} & \multicolumn{4}{c||}{coverage rate @k\%} & \multicolumn{4}{c}{success rate @k\%}\\
        \cline{2-17}
        & 10 & 30 & 50 & 100 & 10 & 30 & 50 & 100 & 10 & 30 & 50 & 100 & 10 & 30 & 50 & 100 \\
        \hline
        GPNet* & 27.9 & 28.9 &	30.1 & 21.9 & 54.0 & 53.7 &	52.2 &	37.7 & 3.0 & 8.0 & 14.3 & 18.6  & 28.9 & 29.2 & 27.2 & 20.8\\
        L2G  & 44.6 &	42.8 &	42.1 &	39.0 & 75.8	& 76.6	& 76.3	& 73.0	&	16.5 &	23.6 &	27.2 &	33.3 &	\textbf{45.0} &	\textbf{44.6} &	43.8 &	41.2\\
        \hline
        GPNet*+\deco \cite{Alliegro_2021_CVPR} & \textbf{54.9} &	48.7 &	42.6 &	32.7 & \textbf{80.0} & 72.7 & 63.5 &	51.0 &	7.5 &	15.8 &	20.5 &	26.2 & 40.3 & 39.2 & 38.0 & 34.1\\
        L2G+DeCo \cite{Alliegro_2021_CVPR} & 52.5 &	\textbf{53.4} & \textbf{51.6} &	\textbf{46.3} & 77.7 & \textbf{78.7} & \textbf{78.4} & \textbf{76.5} & \textbf{17.3} & \textbf{23.6} &	\textbf{27.5} &	\textbf{33.8} & 43.6 &	44.0 &	\textbf{43.9} & \textbf{42.2}\\
        \hline
    \end{tabular}
    \vspace{-3mm}
}
\end{table*}

\begin{figure*}[tb]
    \includegraphics[width=0.99\textwidth]{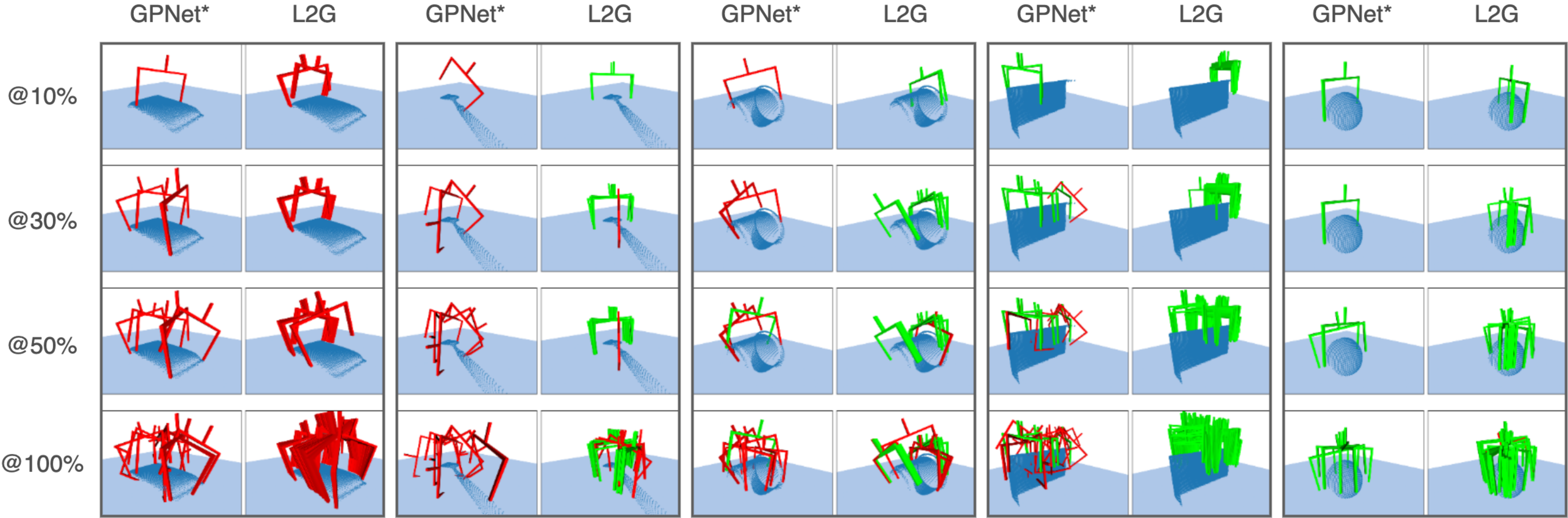}
    \caption{Visualization of the predicted grasps of GPNet$^*$ and L2G for five different objects     from YCB-76 (from left to right: sponge, spoon, cup, sugar box, peach).
        Based on the outcome of the simulation, we color-coded successful grasps in green and unsuccessful ones in red.
        From top to bottom, we increase the parameter~$k$, i.e. the top row contains only the $10\%$ highest-ranked grasp predictions whereas the bottom row contains all grasp predictions.
        }
    \label{fig:qualitative}\vspace{-3mm}
\end{figure*}

On YCB-8, L2G outperforms the GPNet baseline by a large margin in terms of both success rates and coverage. L2G+DeCo shows a further advantage over L2G which is particularly evident on the simulation based results.
Interestingly, the DeCo encoder provides a significant improvement also to GPNet, with GPNet+DeCo reaching similar or even better success rate in simulation than L2G. Still, the results of GPNet+DeCo remain lower than L2G+DeCo in most of the settings.
On the large YCB-76, L2G shows top simulation-based results, demonstrating its generalization abilities. In this case the advantage obtained by using DeCo appears minimal over the already high success rate of L2G. On the other hand, the results of GPNet+DeCo are significantly higher than those of GPNet, despite remaining lower than L2G. 
These findings suggest that the choice of the feature encoder is important for improving generalization on weaker models.
In Fig.~\ref{fig:qualitative} we also present a qualitative analysis to compare the grasp predictions of GPNet and L2G in simulation.
The sponge can be considered an adversarial object: it is {completely} flat and its side length is close to the gripper opening width, hence imposing a high risk of the gripper colliding with either the ground or the object and neither method accomplishes to predict successful grasps.
On the other hand, the peach can be grasped without failure, although there weren't spherical objects in the training set.
Especially with increasing~@k\%, GPNet tends to predict many spurious, unreasonable grasps, which is not the case for L2G.
For elongated objects like the spoon, the grid-based approach of GPNet fails entirely.
L2G, in contrast, is more robust to unknown shapes since it does not rely on any geometric heuristics for the grasp proposal.

\begin{figure}[tb]
    \centering
    \begin{tabular}{@{}c@{~}c@{~}c}
        \includegraphics[height=0.135\textwidth]{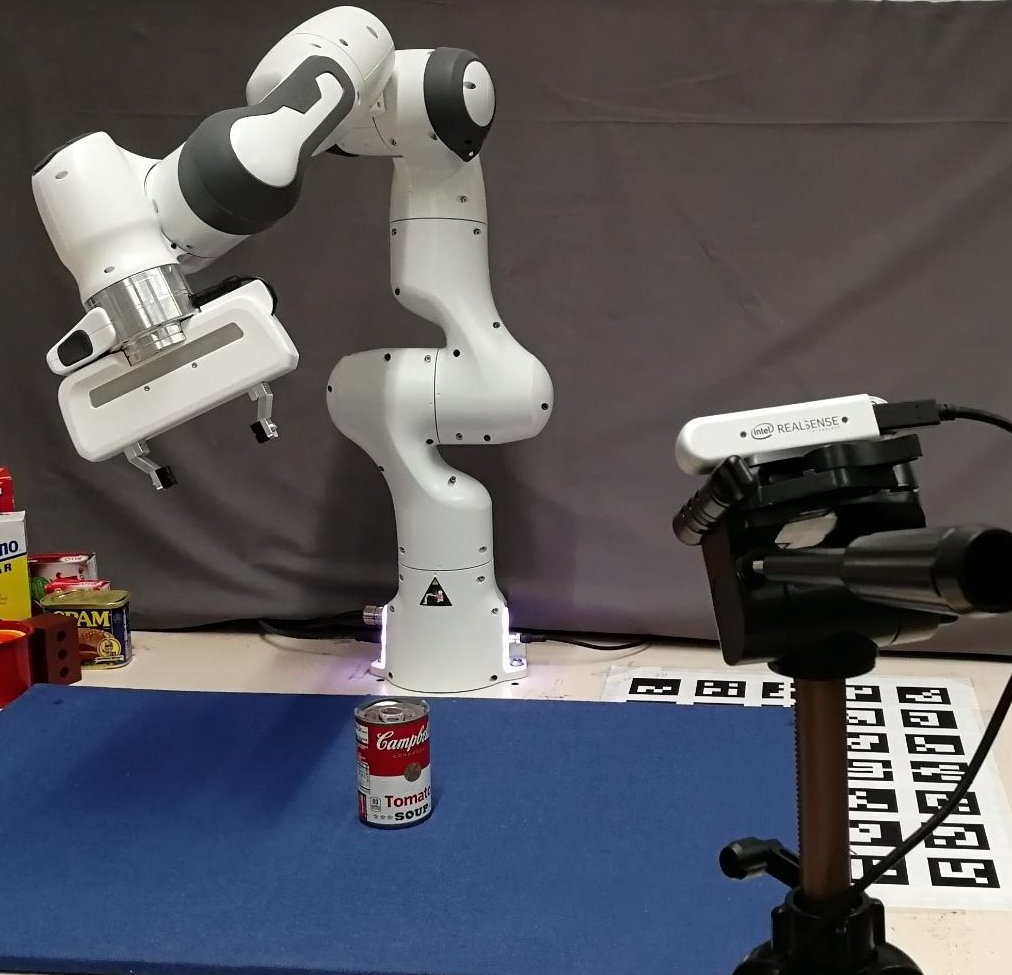} &
        \includegraphics[height=0.135\textwidth]{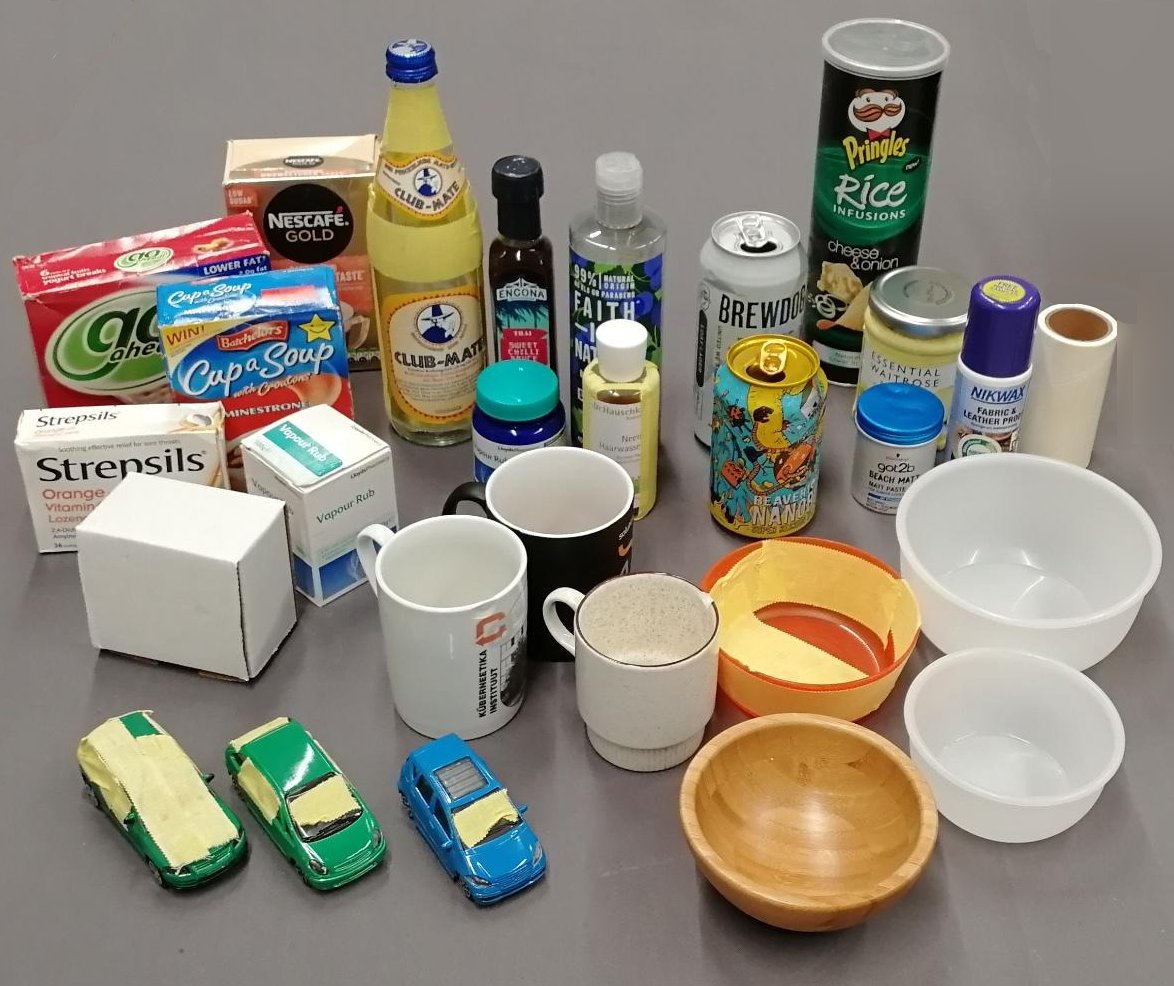} &
        \includegraphics[height=0.135\textwidth]{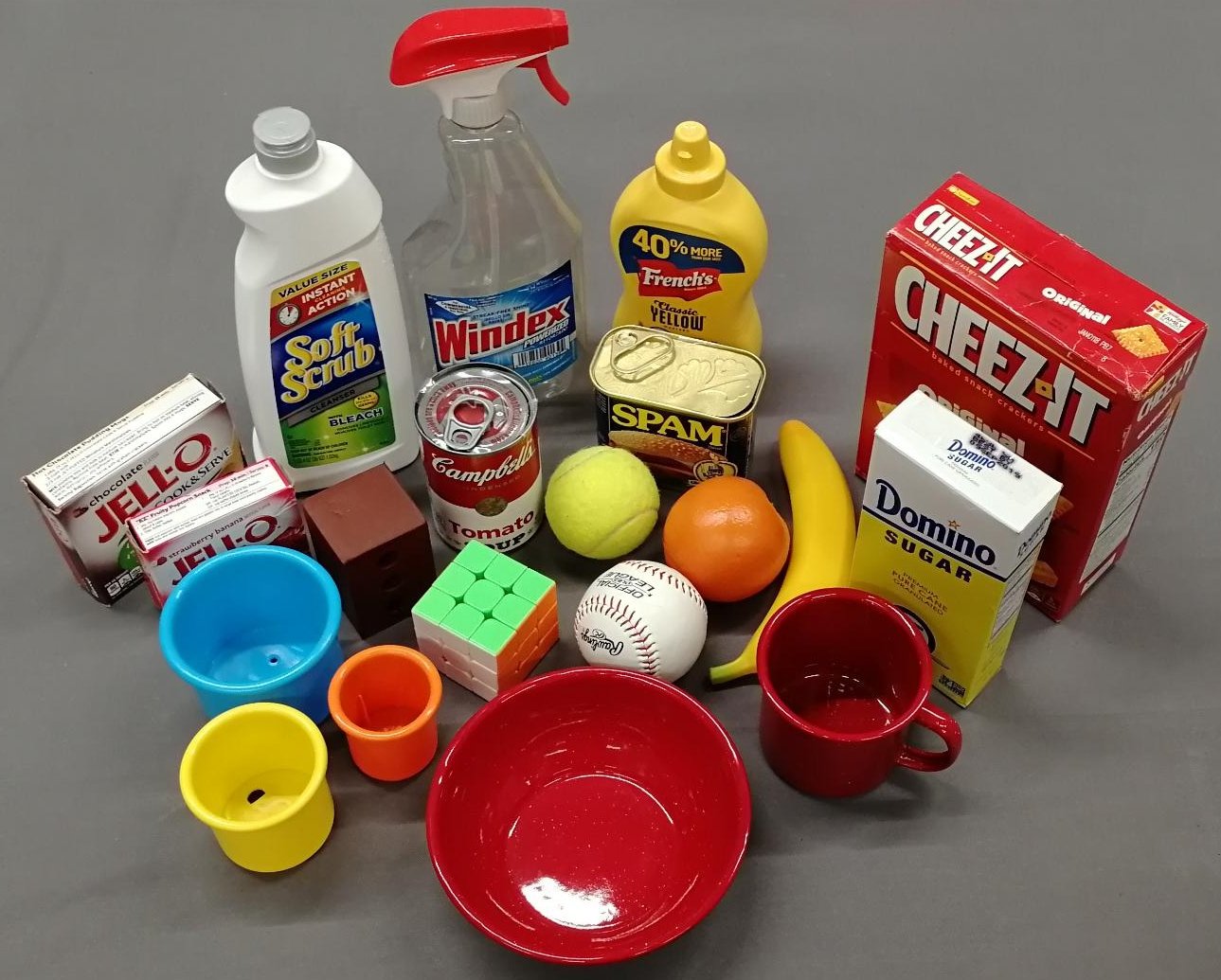}
    \end{tabular}
    \caption{%
    Left: Our real experiments setup with Franka Panda robot and Realsense D415.
    Center: Custom set with 28 objects from similar categories as in ShapeNetSem-8 (from left to right and top to bottom: 6~boxes, 5~bottles, 2~soda cans, 5~cylinders, 3~toy cars, 3~mugs, 4~bowls).
    Right: Selection of 20 YCB objects including shapes from unseen categories.}
    \label{fig:real_exp_objects}
    \vspace{-6mm}
\end{figure}
\begin{table}[t]
    \centering
    \caption{Real-world robot experiments: fraction of successful trials out of five performed on each instance and averaged per category.
    Number of instances per category in parenthesis.
    }
    \label{tab:real_exp}
    \resizebox{0.5\textwidth}{!}{
        \begin{tabular}{c | c|c}
            \hline
            \multicolumn{3}{c}{28 custom items}\\
            \hline
            Category & GPNet$^*$ & L2G \\
            \hline
            box (6) & 0.47 & \textbf{0.73} \\
            soda can (2) & 0.80 & \textbf{1.00} \\
            cylinder (5) & 0.52 & \textbf{0.76} \\
            bottle (5) & 0.25 & \textbf{0.48} \\
            mug (3) & 0.27 & \textbf{0.40} \\
            bowl (4) & \textbf{0.15} & 0.10 \\
            toy car (3) & \textbf{0.33} & 0.27 \\
            \hline
            average & 0.40 & \textbf{0.53}  \\
            \hline
        \end{tabular}
        \renewcommand\arraystretch{1.119}
        \begin{tabular}{c | c|c}
            \hline
            \multicolumn{3}{c}{20 YCB objects}\\
            \hline
            Category & GPNet$^*$ & L2G \\
            \hline
            box (7) & 0.31 & \textbf{0.71} \\
            mug (1) & \textbf{0.20} & \textbf{0.20} \\
            bowl (1) & 0.00 & 0.00 \\
            cylinder (5) & \textbf{0.72} & 0.64 \\
            sphere (3) & \textbf{1.00} & \textbf{1.00} \\
            bottle (3) & 0.07 & \textbf{0.27} \\
            \hline
            average & 0.38 & \textbf{0.47} \\
            \hline
        \end{tabular}
    }\vspace{-4mm}
\end{table}

\subsection{Robot Experiments}
To evaluate the performance in real-world scenarios, we conducted experiments with a Franka Emika Panda robot and an Intel RealSense D415 sensor (see Fig.~\ref{fig:real_exp_objects}).
Based on the highest-ranked grasp prediction obtained from the models (trained purely on synthetic data of ShapeNetSem-8), we plan a trajectory using MoveIt~\cite{MoveIt}.
If no feasible plan can be found, we swap the contact points and plan for this symmetric grasp.
If this still does not give a feasible trajectory, we resort to the second or at most the third grasp prediction.
The trial counts as successful only if the object is continuously in contact with both fingertips from grasp execution until release.

We performed five trials for each object in two sets: one with 28~custom items from the same categories as in ShapeNetSem-8, and one with 20~YCB objects for better experimental reproducibility (see Fig.~\ref{fig:real_exp_objects}).
The results are displayed in Table~\ref{tab:real_exp} and indicate that the grasping performance varies strongly in relation to the object category. 
The bowls, which have been grasped with least success, required different modes of grasping:
they can be grasped around the circumference only if the diameter is smaller than the gripper opening width, else they must be grasped along the rim.
Both GPNet and L2G did not cope well with this mode switch.
On the other hand, sphere-like objects and soda cans could be grasped by L2G without failure.
Across categories, we observed that both grasping approaches are sensitive to object size.
In ShapeNetSem-8, all objects are scaled to have their smallest dimension $>$60mm and their largest $<$150mm. Based on this criterion we can separate the 28 custom items into two sub-groups: one with the 10 objects which fit the size constraints, the other with the remaining 18 which do not. It is genuinely hard to find real objects from the categories bottle, toy car, bowl, and cylinder with these dimensions and almost all the instances of those categories are in the second group.
As could be expected the results on the 10 object group (GPNet: 0.44, L2G: 0.62) are higher than those on the 18 object group (GPNet: 0.36, L2G: 0.49), but the improvement of L2G over GPNet remains confirmed.

A further challenge originates from the simulation to reality gap to which also the gripper type contributes.
The gripper used in the simulation environment was a Robotiq-2F85 gripper, whereas we used a Panda gripper with shorter finger length and slightly smaller opening width.
This is crucial when grasping (flat) objects close to the ground, like the toy cars.
Overall, L2G outperforms GPNet by a significant margin and is also ahead in most category-specific comparisons, with the biggest margins for boxes and bottles.

\section{CONCLUSIONS}
In this paper we introduced L2G, our end-to-end method for 6-DOF grasping from partial object point clouds which leverages a differentiable sampling strategy to identify the visible contact points, and a feature encoder which combines local and global cues.
Overall, L2G is guided by a multi-task objective to produce a diverse set of grasps by optimizing contact point sampling, grasp regression, and grasp classification.
In our experimental analysis we thoroughly compared L2G to the main competitor GPNet~\cite{WuChenNeurIPS20} and could demonstrate its advantages: 
L2G predicts a larger, more diverse set of reliable grasps.
Furthermore, it better generalizes to unseen objects with significant shape variations.
Using the pre-trained feature encoder DeCo~\cite{Alliegro_2021_CVPR} instead of a standard PointNet++ significantly boosts generalization performance.

Here we considered single object scenes to maintain the main focus on the robustness and effectiveness of the proposed approach. Still, by its design, L2G can be easily adapted for scenes containing multiple objects. 
The point sampling procedure could also incorporate prior knowledge about the downstream manipulation task when available.
We plan to investigate both these directions in future work.

\normalem

\bibliographystyle{IEEEtran}
\bibliography{root}

\end{document}